\def\year{2021}\relax
\documentclass[letterpaper]{article} 
\usepackage{aaai21}  
\usepackage{times}  
\usepackage{helvet} 
\usepackage{courier}  
\usepackage[hyphens]{url}  
\usepackage{graphicx} 
\usepackage{natbib}  
\usepackage{caption} 
\usepackage{times}
\usepackage{latexsym}
\usepackage{url}
\usepackage{float}
\usepackage{amsmath}
\usepackage{amssymb}
\usepackage{microtype}
\usepackage{subfigure}
\usepackage{booktabs}
\usepackage{amsthm}
\usepackage{amsfonts}
\usepackage{arydshln}
\usepackage{multirow}
\usepackage{multicol}
\usepackage{comment}

\usepackage{tcolorbox}

\def\KL{\textsf{KL}} 

\usepackage{verbatim}

\usepackage{anyfontsize}

\usepackage{color}
\usepackage{tikz}
\usetikzlibrary{arrows,shapes,decorations,automata,backgrounds,fit,petri}
\usepackage{adjustbox}

\newcommand{\RN}[1]{%
	\textup{\lowercase\expandafter{\it \romannumeral#1}}%
}

\makeatletter
\newcommand{\distas}[1]{\mathbin{\overset{#1}{\kern\z@\sim}}}%

\usepackage{enumitem}

\newcommand{\beq}{\vspace{0mm}\begin{equation}}
\newcommand{\eeq}{\vspace{0mm}\end{equation}}
\newcommand{\beqs}{\vspace{0mm}\begin{eqnarray}}
\newcommand{\eeqs}{\vspace{0mm}\end{eqnarray}}
\newcommand{\barr}{\begin{array}}
\newcommand{\earr}{\end{array}}

\newcommand{\xv}{\boldsymbol{x}}
\newcommand{\yv}{\boldsymbol{y}}
\newcommand{\zv}{\boldsymbol{z}}

\newcommand{\thetav}{\boldsymbol{\theta}}

\newcommand{\phiv}{\boldsymbol{\phi}}

\newcommand{\Dcal}{\mathcal{D}}

\usepackage{algorithm}
\usepackage{algorithmic}

\usepackage{tabularx}
\usepackage{graphicx}
\usepackage{adjustbox}
\usepackage{makecell}

\usepackage{ragged2e}

\title{Transformer-based Conditional Variational Autoencoder \\ for Controllable Story Generation}

\author{Le Fang$^\dagger$,
    ~~Tao Zeng$^\mathsection$,
    ~~Chaochun Liu$^\mathsection$, 
    ~~Liefeng Bo$^\mathsection$, 
    ~~Wen Dong$^\dagger$,
    ~~Changyou Chen$^\dagger$\\
    \textnormal{ $^\dagger$University at Buffalo,~~
    $^\mathsection$JD Finance America Corporation, AI Lab} \\
    \textnormal{ \{lefang, wendong, changyou\}@buffalo.edu }\\ 
    \textnormal{ \{tao.zeng, chaochun.liu, liefeng.bo\}@jd.com}
 }

\date{}

\begin{document}
\maketitle

\begin{abstract}
We investigate large-scale latent variable models (LVMs) for neural story generation---an under-explored application for open-domain long text---with objectives in two threads: generation effectiveness and controllability. LVMs, especially the variational autoencoder (VAE), have achieved both effective and controllable generation through exploiting flexible distributional latent representations. Recently, Transformers and its variants have achieved remarkable effectiveness without explicit latent representation learning, thus lack satisfying controllability in generation. In this paper, we advocate to revive latent variable modeling, essentially the power of representation learning, in the era of Transformers to enhance controllability without hurting state-of-the-art generation effectiveness. Specifically, we integrate latent representation vectors with a Transformer-based pre-trained architecture to build conditional variational autoencoder (CVAE). Model components such as encoder, decoder and the variational posterior are all built on top of pre-trained language models---GPT2 specifically in this paper. Experiments demonstrate state-of-the-art conditional generation ability of our model, as well as its excellent representation learning capability and controllability.
\end{abstract}

\section{Introduction}
Neural text generation has achieved remarkable success to enable both effective and controllable generation at a level that computational models can write like humans to satisfy practical needs. Among various research objectives, the most significant ones are the effectiveness and the controllability of generation, where there are always emerging opportunities and challenges.

Deep latent variable models (LVMs), especially variational autoencoder (VAE) \cite{kingma2013auto, rezende2014stochastic} have been a significant class of methods to achieve both effective and controllable generation \cite{bowman2015generating, miao2016neural, zhao2017learning, zhao2018unsupervised, zhou2017multi, hu2017toward, bao2019generating, shah_generative}. These models generally work with recurrent neural networks (RNN) such as Long short-term memory (LSTM) \cite{hochreiter1997long} and Gated recurrent unit networks (GRU) \cite{cho2014learning}. The advantage of LVMs is to learn and exploit flexible distributional latent representations to capture holistic features of input and further guide the generation of sentences. Such powerful representation learning can deal with both the effectiveness and the controllability of generation.

In recent years, Transformers \cite{vaswani2017attention} and its variants have become the main-stream workhorses and boosted previous generation effectiveness by large margins. Models based on similar self-attention architectures \cite{devlin2018bert, radford2018improving, radford2019language} could leverage both big models and big training data. A dominant paradigm emerges to be ``pre-training + fine-tuning'' on a number of natural language processing tasks. Even without explicitly learning latent representations, Transformer-based models could effectively learn from training data and generate high-quality text. It's thrilling to witness computational models generate consistent long text in thousands of words with ease. However, given state-of-the-art generation effectiveness, controllability of these models---especially when generating long text---is still under-explored. The emerging challenge is, how to achieve controllable generation in the era of Transformers and a long text setting?

In this paper, we advocate to revive latent variable modeling, essentially the power of representation learning, in the era of Transformers to enhance controllability without hurting state-of-the-art generation effectiveness. Specifically, we integrate latent representation vectors with self-attention based pre-trained architecture to build a conditional variational autoencoder (CVAE). Model components such as encoder, decoder and the variational posterior are all built on top of pre-trained language models---GPT2 \cite{radford2019language} specifically. We demonstrate excellent representation learning capability and controllability of our Transformer-based LVMs through learning and manipulating the latent representation vectors.

On the application side, we emphasize a much challenging and under-explored task, i.e. neural story generation, which creatively writes open-domain long text in hundreds or thousands of words conditioned on very short and abstract prompts \cite{fan2018hierarchical}. The task featured with much longer output leads to higher complexity and more flexibility in a broader space than short text generation. Previous literature \cite{fan2018hierarchical, mao2019improving, see2019massively, ziegler2019encoder} have at most studied how to effectively learn the mapping between prompt and story through explicit end to end (end2end) training. However, controllability in such a setting has rarely been studied. For instance, how to control story development and semantic transition during the spanning of long text? Pure end2end learning seems quite rigid, which could miss flexible and controllable mechanisms inside a black box. A reasonable solution for this issue is to introduce latent representation vectors, which is the treatment we consider in this paper.

To summarize, our paper is among the first works, by our knowledge, to build Transformer-based latent variable models to solve the controllability issue in the setting of long text generation. Recently, we notice an independent parallel work \cite{li2020optimus}, which proposes similar Transformer-based architecture to incorporate latent representation vectors. We note that there are a number of differences between this work and ours. Most significantly, we considered both VAE and CVAE in a long text setting, while \citet{li2020optimus} considered a pre-trained VAE model in traditional short text setting. Our datasets and source code is available on GitHub\footnote{\url{https://github.com/fangleai/TransformerCVAE}}.

\section{The Model Architecture}
\subsection{Conditional Variational Autoencoder}
Conditional story generation \cite{fan2018hierarchical} refers to generating open-domain long text based on a short prompt, which provides either a starting point or an abstract summary for the writing. In this paper, we propose a Transformer-based conditional variational autoencoder to learn the generative process from prompt to story.

\begin{figure}[ht!]
\centering
\includegraphics[scale=0.4]{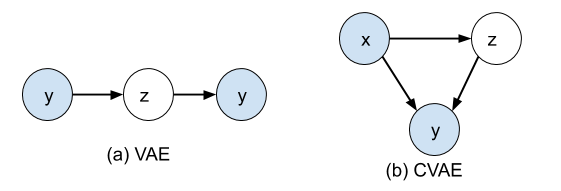}
\vspace{-0mm}
\caption{Graphical Model of VAE and CVAE. In controllable story generation, $\xv$ and $\yv$ refer to a prompt and a story, respectively. $\zv$ refers to a latent variable.}
\label{fig:vae_cvae_2}
\end{figure}

\paragraph{Variational Autoencoder (VAE)}~\cite{bowman2015generating} Figure~\ref{fig:vae_cvae_2} illustrates the graphical model of VAE, an unsupervised learning method for unconditional generation. VAE consists of a generative network (decoder) and an inference network (encoder). Given a language dataset $\mathcal{D}=\{\yv_{i}\}_{i=1}^{\left|\mathcal{D}\right|}$, where $\yv_{i} = [y_{1i}, \cdots, y_{Ti} ]$ represents $i$th sentence of length $T$. With a prior distribution $p(\zv)$, VAE generates a sentence $\yv$ using the deep generative network $p_{\thetav}(\yv|\zv)$ parameterized by $\thetav$. The prior $p(\zv)$ is typically assumed to be a standard multivariate Gaussian. The decoder $p_{\thetav}(\xv|\zv)$ typically takes an auto-regressive form $p_{\thetav}(\xv|\zv)=\prod_{t=1}^{T}p_{\thetav}(x_{t}|x_{<t},\zv)$. In this paper, we will build the decoder based on the pre-trained GPT2 rather than traditional recurrent neural networks. 

The goal of VAE training is to maximize the marginal data log-likelihood $\mathbb{E}_{\yv\sim \mathcal{D}}[\log p_{\thetav}(\yv)]$. However, posterior inference is generally intractable. Consequently, an $\phiv$-parameterized encoder is introduced to approximate $p_{\thetav}(\zv|\yv)\propto p_{\thetav}(\yv|\zv)p(\zv)$ with a variational distribution $q_{\phiv}(\zv|\yv)$. Variational inference is employed for VAE learning, yielding the following evidence lower bound (ELBO):
\begin{align}
 & \mathbb{E}_{\xv \sim \Dcal} \text{log} p_{\thetav}(\xv) \geq \nonumber \\ 
 & \mathbb{E}_{\xv\sim \Dcal}\left[\mathbb{E}_{\zv\sim q_{\phiv}(\zv|\xv)}\text{log}p_{\thetav}(\xv|\zv) \right] \nonumber \\
  & - \mathbb{E}_{\xv\sim \Dcal}\left[\KL\left(q_{\phiv}(\zv|\xv)\parallel p(\zv)\right)\right] \label{eq:ELBO-vae}
\end{align}

\paragraph{Conditional Variational Autoencoder (CVAE)}~\cite{zhao2017learning} Figure~\ref{fig:vae_cvae_2} also illustrates the CVAE, an adaptation of VAE to fit supervised learning and conditional generation. Given a training dataset of pairs $\mathcal{D}=\{\xv_{i}, \yv_{i}\}_{i=1}^{\left|\mathcal{D}\right|}$, where $\xv_{i} = [x_{1i}, \cdots, x_{Ti} ], \yv_{i} = [y_{1i}, \cdots, y_{Ti} ]$ represents $i$th sentence of length $T$. In controllable story generation, $\xv$ and $\yv$ refer to a prompt and a story, respectively. Given an input $\xv$, CVAE encodes the prior knowledge of latent code as $p(\zv|\xv)$, and generates target $\yv$ using the deep generative network $p_{\thetav}(\yv|\xv, \zv)$ parameterized by $\thetav$.

The goal of CVAE is to maximize the conditional data log-likelihood $\mathbb{E}_{\xv, \yv\sim \mathcal{D}}[\log p_{\thetav}(\yv|\xv)]$. Similarly, variational inference is employed for CVAE learning, yielding the following evidence lower bound (ELBO):
\begin{align}
 & \mathbb{E}_{\xv, \yv \sim \Dcal} \text{log} p_{\thetav}(\yv|\xv) \geq \nonumber \\
 & \mathbb{E}_{\xv, \yv\sim \Dcal}\left[\mathbb{E}_{\zv\sim q_{\phiv}(\zv|\xv, \yv)}\text{log}p_{\thetav}(\yv|\xv, \zv) \right] \nonumber \\
 & - \mathbb{E}_{\xv, \yv\sim \Dcal}\left[\KL\left(q_{\phiv}(\zv|\xv, \yv)\parallel p(\zv|\xv)\right)\right] \label{eq:ELBO-cvae}
\end{align}
Note that both the prior $p_{\thetav}(\zv|\xv)$ and posterior $q_{\phiv}(\zv|\xv, \yv)$ are learnable in CVAE.

\subsection{Architecture Design}
\begin{figure*}[th!]
\centering
\includegraphics[width=14cm]{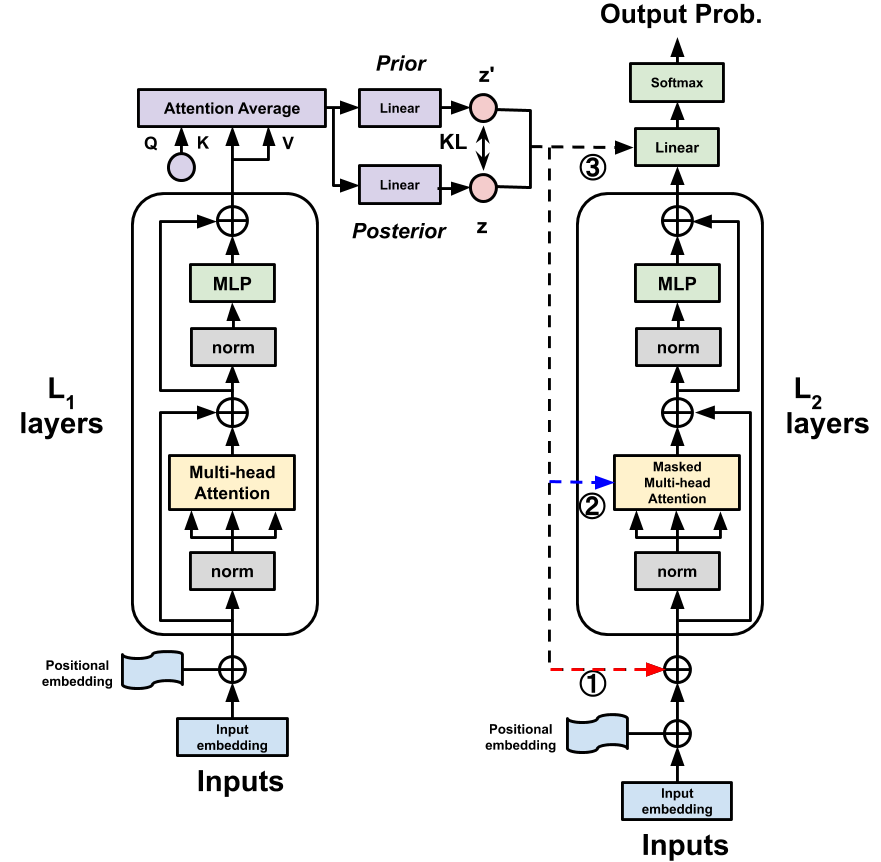}
\vspace{-0mm}
\caption{Our CVAE model architecture. Note that we use CVAE \textcircled{1}, CVAE \textcircled{2}, CVAE \textcircled{3} to represent model variants that use the \textcircled{1}, \textcircled{2}, \textcircled{3} way of latent code injection respectively. The final model are not meant to use all ways but only subset combinations as implied by performance.}
\label{fig:tcvae}
\end{figure*}

Our model architecture is illustrated in Figure~\ref{fig:tcvae}. Basically, it consists of a prior, posterior and conditional generator based on multi-layer self-attention architecture \cite{vaswani2017attention}, more specifically on top of pre-trained models.

\paragraph{Parameter initialization with GPT2}
In order to exploit the power of pre-trained models, we propose to reuse the GPT2 model \cite{radford2019language} as our decoder. For ease of computation, we adopt the smallest public version with $L=12$ layers, $H=12$ heads per layer, model dimension of $d=768$ units and total parameters of 117M. The encoder has $L_1$=6 unmasked/bi-directional self-attention layers, whose parameters are initialized to the parameters of the first $L_1$ layers of the GPT2 model (initialized but not shared afterwards). Moreover, the word embedding and positional embedding tables in the encoder and decoder are shared. 

Comparing with masked/uni-directional structure in decoder for auto-regressive generation, the key point is to have unmasked/bi-directional structure in the encoder to allow full information scope. In this sense, our design is comparable with \cite{li2020optimus} to reuse BERT in the encoder and GPT2 in the decoder. However, we avocate two main design differences: 1) We note that BERT uses word piece (WPE) embeddings for tokenization and GPT-2 uses Byte Pair Encoding (BPE), leading to totally different vocabulary books. \cite{li2020optimus} resorts to keeping both tokenizations for all inputs and outputs; while our design has none of such issues. 2) \cite{li2020optimus} only works with short sentences typically less that 64 words, while our model works with hundreds or thousands of words in a minimal run. In our case, a model with a full layer encoder ($L_1$=12) is empirically too large to fit a single GPU memory. In order to save memory and considering that the first several layers of GPT2 may implicitly serve to encode features, our model only use $L_1$=6 layers in the encoder\footnote{Our experiment confirms that using full layers in encoder has limited improvement in performance comparing to using $L_1$=6 layers in encoder.}.

\begin{figure}[t!]
\centering
\includegraphics[scale=0.13]{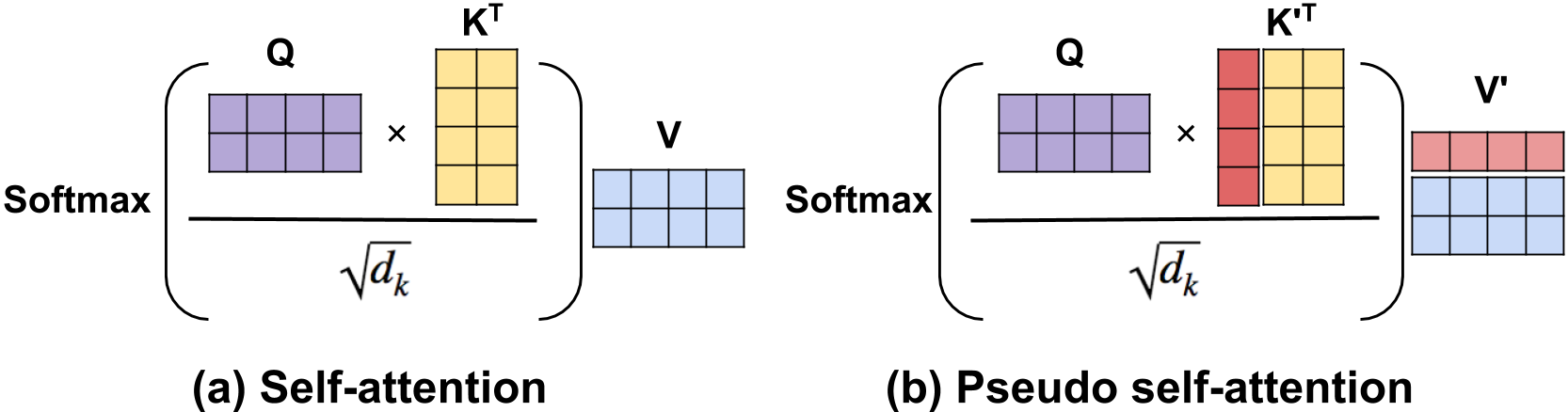}
\vspace{-0mm}
\caption{Self-attention (SA) and pseudo self-attention (PSA). SA is widely used in our model including the attention-average block, when pseudo self-attention can be used as the \textcircled{2} way to feed latent code (key and value colored in light red) to the decoder.}
\label{fig:sa_psa}
\end{figure}

\paragraph{Latent code from the encoder}
Traditional RNN/LSTM encoders typically only use the last hidden state from the encoder to produce a latent space. This is insufﬁcient to summarize sequential data and keep long-term knowledge. In our model, representations from self-attention layers are a sequence of vectors with total number equal to the number of input tokens. To utilize all the information, we define an attention-average block right afterwards to merge variable length sequence of vectors into a single vector. The attention average block essentially perform a multi-head self-attention as Figure~\ref{fig:sa_psa}(a) using a learnable single query $Q=q_{avg}\in\mathbb{R}^{d}$, and $K=V$ taken as the variable length sequence of vectors from the last blocked self-attention layer. The single vector representation is then passed to linear layers to predict prior and posterior distribution, respectively.

In terms of model components, we define both the prior and the variational posterior as isotropic Gaussian distributions, {\it i.e.}, $N(\mu,\sigma^{2}I)$, with learnable mean vector and ``$\log\sigma$'' vector. The KL divergence between the prior and posterior in Eq. \ref{eq:ELBO-cvae} is therefore analytically solvable. Traditional reparameterization trick \cite{kingma2013auto} is used to allow gradient passing through Gaussian sampling. Note that in our model, the prior distribution $p(\zv|\xv)$ and variational posterior distribution $q_{\phiv}(\zv|\xv, \yv)$ share all the parameters except the linear layers predicting their mean and variances to promote prior posterior alignment. 

In contrast to Transformers \cite{vaswani2017attention} and other models that learn sequence of encoding vectors, our model is dedicated to learning a single vector as an explicit latent representation.

\paragraph{Feeding latent code to the decoder}
With a single latent code representation\footnote{Latent code can have dimension $d'\neq d$, in which case linear projection layers are needed before feeding latent code to the decoder to ensure identical dimensions.} $z\in\mathbb{R}^{d}$ and a ``GPT2'' decoder, we investigate three mainstream ways of latent code injection inspired by previous literatures \cite{cheng2019variational, ziegler2019encoder, wang2019t}.  
\begin{itemize}
    \item[\textcircled{1}] \textsc{Input:} $z$ is added to each input token during decoding, {\it i.e.}, added with word embeddings and positional embeddings element-wisely.
    \item[\textcircled{2}] \textsc{PSA\footnote{In recent literature \cite{li2020optimus}, they also studied a way of latent injection described as ``Memory'' vector. Essentially, ``Memory'' is identical or equivalent to our ``PSA''. }:} inject latent code $z$ in a per-layer basis. Specifically, we first project $z\in\mathbb{R}^{d}$ into $z_L\in\mathbb{R}^{d\times L}$ through a linear layer, so that it can be split into $L$ vectors $[z_1, \cdots, z_L]$, with $z_l$ being fed into the $l$-th blocked self-attention layer. As presented in \cite{ziegler2019encoder} and shown in  Figure~\ref{fig:sa_psa}(b), pseudo self-attention could absorb extra encoding embeddings into a pre-trained GPT2 self-attention structure through
\begin{align}
   \text{PSA}(Q, K', V') = \text{softmax}(\frac{QK'^{T}}{\sqrt{d_{k}}})V' \label{eq:psa_dot_prod2}
\end{align}
where $Q, K, V \in\mathbb{R}^{l\times d}$ are the original input embeddings participating self-attention; $K'=\dbinom{z_K}{K}\in\mathbb{R}^{(1+l)\times d}$, $V'=\dbinom{z_V}{V}\in\mathbb{R}^{(1+l)\times d}$ are augmented key and value matrices with projected latent code $z_K$, $z_V$ from $z$ filling the first row; $\dbinom{\cdot}{\cdot}$ means concatenation by rows. Here, we abbreviate per-layer code $z_l$ to $z$ for notation simplicity.
    \item[\textcircled{3}] \textsc{Softmax:} in the original GPT2, an embedding vector $h\in\mathbb{R}^{d}$ from the last blocked attention layer is projected to a pre-softmax logit vector $p\in\mathbb{R}^{V}$ through a linear head, where $V$ is the vocabulary size used in tokenization. When a latent code should be injected in such a position, a new and shared linear head will be initialized and learned in order to project $z\in\mathbb{R}^{d}$ into $p_z\in\mathbb{R}^{V}$. Finally we send $p+p_z$ for the softmax and output.
\end{itemize}

We empirically study all the three ways of latent code injection into the decoder, and present comparison in the experiment section. 

\subsection{Training and Generation}
We train our CVAE model according to the negative loss objective in \eqref{eq:ELBO-cvae}. For conditional story generation, the input to the prior distribution $p(\zv|\xv)$ is purely the prompt and the input to posterior distribution $q_{\phiv}(\zv|\xv, \yv)$ is the connected sequence of prompt and story split by a special token `$<\!\!|\text{endoftext}|\!\!>$''. The conditional generative distribution $p_{\thetav}(\yv|\xv, \zv)$ is implemented as decoding with a text prefix ``prompt + $<\!\!|\text{endoftext}|\!\!>$'' and feeding the latent code.

To avoid learning deviation caused by random initialized parameters, we freeze pre-trained parameters initialized from GPT2 in the first several iterations of training, i.e. 10K iterations, and unfreeze them afterwards. 

To alleviate the notorious posterior collapse issue, we take a cyclic annealing schedule \cite{fu2019cyclical} by adjusting the coefficient $\beta$ before KL divergence in \eqref{eq:ELBO-cvae}. Specifically, we have kept $\beta$ close to zero in the first half of cyclic schedule, linearly annealed $\beta$ to 1 in the next one fourth of cyclic schedule and kept $\beta=1$ in the remaining one fourth of cyclic schedule. The purpose of such schedule is to exploit the period that $\beta$ is close to zero, which pushes the model towards a pure autoencoder. Note that autoencoder \cite{bourlard1988auto} learns point estimate of latent code instead of distributional representation to generate target data, which could improve generation effectiveness.

During generation, a short prompt text is fed to the encoder, and a latent code is sampled from the prior distribution to guide the decoding. This procedure is the same as how traditional CVAE works.

\section{Related Work}\label{sec:related}
\subsection{Controllable Story Generation}
Most of previous works on text generation consider a setting of short text. For controllable generation, they mainly consider certain global aspects of text, with the most common aspects being sentiment and topic \cite{shen2017style,zhao2018adversarially,hu2017toward,fang2019implicit,dathathri2019plug,keskar2019ctrl,li2020optimus, wang2020faithful}. Researchers have attempted short story generation with fine-grained control through plots, plans or the so-called storylines \cite{peng2018towards,yao2019plan}, leading to a wide usage and benchmark on 5-lines story dataset $\mathtt{ROCStories}$ \cite{mostafazadeh2016corpus}. 

In recent years, \cite{fan2018hierarchical} proposes story generation as a test bed of open-domain long text generation \cite{fang2021outline}. \cite{ziegler2019encoder} initiates the research of conditionally generating story based on a pre-trained GPT2.

Though achieving promising results, very few works have been presented to improve controllability in the setting of long text. This work is, by our knowledge, the first work to build a Transformers-based latent variable model to improve controllable open-domain long text generation.

\subsection{Transformers-based LVMs}
Recently, there are several works building latent variable models on top of the Transformer. One main class of work is conditional VAEs with Transformer-based components in non-autoregressive sequence generation, especially non-autoregressive machine translation \cite{shu2020latent,ma2019flowseq,kasai2020parallel,han2020non}. Another class of work is on dialogue generation with conditional VAEs \cite{bao2019plato, lin2020variational} for response diversity.

From another perspective, \cite{wang2019t} aims to learn latent representations for story completion, where latent variables are only used to fill very short text blank. \cite{cheng2019variational} proposes a semi-supervised method using a Transformer-based VAE to solve aspect-term sentiment analysis problem. The method also disentangles latent space for aspect-specific sentiment and the lexical context, respectively. A recently released literature \cite{li2020optimus} proposes large-scale VAE as a pre-trained model. The model is first pre-trained with massive unlabelled corpus and then fine-tuned in various down-stream generation tasks that traditional RNN/LSTM based VAEs have attempted.

Our paper indeed gets some inspirations from previous works of Transformer-based latent variable models on architecture side. However, our model is motivated to enhance controllability of generation and deals with a challenging setting of long-text generation.

\section{Experimental Results and Discussions}\label{sec:tvae-exp}
\subsection{Pre-Experiment on VAE Architecture}
In order to verify our idea of Transformer-based latent variable models, we first conduct a pre-experiment with the VAE architecture on two small datasets. The VAE implemented is a simpler version of our CVAE model shown in Figure~\ref{fig:tcvae}, where the prior is defined as standard spherical Gaussian $N(1,I)$. Moreover, the VAE conducts pure unsupervised learning, where unlabelled language texts are encoded into a distributional representation space and then decoded back. 

\paragraph{Datasets:} The two relatively small datasets are introduced in the following for VAE learning respectively with statistics shown in Table~\ref{table:datasets-small}.

The $\mathtt{Arxiv}$ is an online dataset \cite{cunha_arxiv} that extracts abstracts from ``arxiv'' articles. Specifically, a topic query is searched among arxiv abstracts and the matched ones are collected. The three topic queries we used are ``artificial intelligence'', ``computer vision'' and ``language generation'', leading to around 12K article abstracts respectively having such topic words. The $\mathtt{Yelp}$ is a public dataset \cite{yang2017improved, he2018lagging} with restaurant reviews collected from the ``Yelp'' website. Reviews are associated with user ratings from one to five stars. We binarize reviews with user rating above three as positive, and below three as negative, leading to a binary sentiment dataset.


\begin{table}[t]
\begin{centering}
\begin{adjustbox}{scale={1.0}{1.0},center}
\begin{tabular}{c|c|c|c}
\toprule
Dataset & Num. of text & Split & Avg. length \tabularnewline
\midrule
$\mathtt{Arxiv}$ & 35228 & 90-5-5 & 233 \tabularnewline
$\mathtt{Yelp}$ & 120 K & 10-1-1 & 96.7  \tabularnewline
\bottomrule
\end{tabular}
\end{adjustbox}
\end{centering}
\caption{Statistics of datasets in VAE pre-experiment.}
\label{table:datasets-small}
\end{table}

\begin{figure}[t!]
\centering
\subfigure[Arxiv.]{
\label{fig:ax_pre}
\includegraphics[width=4.0cm]{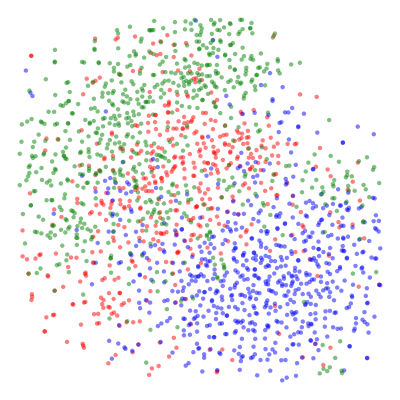}}
\hspace{0.0cm}
\subfigure[Yelp.]{
\label{fig:yp_pre}
\includegraphics[width=4.0cm]{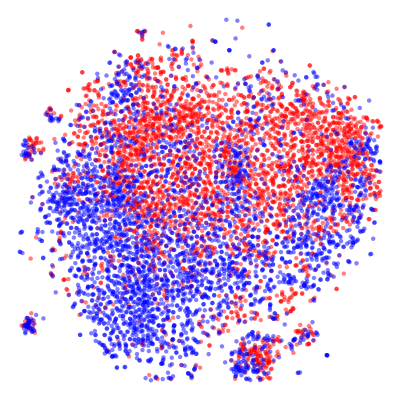}}
\caption{Representation learning of stories in pre-experiment. (a) Arxiv: Topic are draw in different colors: red for artificial intelligence; blue for computer vision; green for language generation; (b) Yelp: Sentiment are draw in two colors: red for negative; blue for positive.}
\label{fig:repre-pre}
\end{figure}

\paragraph{Visualization Results}
Figure~\ref{fig:repre-pre} visualizes the posterior $\zv$ of texts in the test dataset in 2D space using t-SNE~\cite{maaten2008visualizing}. 
As we can see, meaningful latent spaces can be learned, which are able to cluster high-dimension data according to proximity between their latent codes. Interestingly, for the $\mathtt{Arxiv}$ dataset, cluster of ``artificial intelligence'' lies between clusters of ``computer vision'' and ``language generation'', which coincides with our understanding of these topics. Such visualization shows encouraging signs on representation learning power of our model.

\subsection{Experimental Settings}
\paragraph{Datasets:} We conduct conditional story generation on two datasets, $\mathtt{WritingPrompts}$ and $\mathtt{WikiPlots}$, with statistics shown in Table~\ref{table:datasets-tvae}. The datasets are publicly available and meet our target of open-domain long text corpora. We have investigated several other most commonly used public datasets in conditional text generation. Another dataset commonly used in story-plot generation is $\mathtt{ROCStories}$ \cite{mostafazadeh2016corpus,yao2019plan}, which consists of 5-lines stories, thus are too short to use in our task. 

\begin{table}[t]
\begin{centering}
\begin{adjustbox}{scale={1.0}{1.0},center}
\begin{tabular}{c|c|c|c|c|c}
\toprule
Dataset & \makecell{Num.\\ stories} & Split & \makecell{Prompt\\ avg.\\ len.} & \makecell{Story\\ avg.\\ len.} & \makecell{Story\\ avg.\\ paragraphs} \tabularnewline
\midrule
$\mathtt{WP}$ & 303 K & 90-5-5 & 25.4 & 674.5 & 6.3 \tabularnewline
$\mathtt{WI}$ & 113 K & 90-5-5 & 3.4 & 332.9 & 3.1 \tabularnewline
\bottomrule
\end{tabular}
\end{adjustbox}
\end{centering}
\caption{Statistics of datasets for controllable story generation. $\mathtt{WP}$: $\mathtt{WritingPrompts}$; $\mathtt{WI}$: $\mathtt{WikiPlots}$.}
\label{table:datasets-tvae}
\end{table}

For the two datasets adopted,
the $\mathtt{WritingPrompts}$ is a dedicated large scale hierarchical story generation dataset collected from Reddit's ``WritingPromts'' forum \cite{fan2018hierarchical,mao2019improving}. Given a prompt as a rough guide or starting point, stories are multi-paragraph short novels written by human users;
the $\mathtt{WikiPlots}$\footnote{\url{https://github.com/markriedl/WikiPlots}} contains story plot about books, novels, films, and etc, extracted from English language Wikipedia. Each story plot is paired with a short title, which is used similarly as prompt in the $\mathtt{WritingPrompts}$.

\subsection{Benchmark Models}
Each of our benchmark models serves designated purposes. Note that we don't benchmark with other pre-trained language model bases which may be much more powerful than GPT2. We also don't choose some popular controllable text generators such as \cite{hu2017toward,dathathri2019plug} since they either only work in a short text setting or discuss a different notion of control. 

By comparing with a state-of-the-art specialized-architecture task-specific story generation model \cite{fan2018hierarchical}, we evaluate models' in-domain generation performances. Fusion models in \cite{fan2018hierarchical} takes a convolutional seq2seq model structure with a fusion training mechanism. Although similar self-attention architectures are used, the fusion model is still different with our Transformer-based architectures on the design of key and value vectors. 

By comparing with a state-of-the-art transfer learning method based on GPT-2 models---pseudo self attention (PSA) \cite{ziegler2019encoder}, we compare our CVAE model with a pure supervised training method for conditional generation. The pseudo self attention introduces new projection matrices to absorb a sequence of embedding vectors from input to the self-attention computational framework. Note that we may use pseudo self attention (PSA) as one way of latent code injection (\textcircled{2}), but there are key differences: our model only injects a single encoded vector, rather than a sequence of encoding vectors in original PSA; our CVAE model has exploited the notion of distributional representation to learn a representation space to enable flexible controllability. In another words, our CVAE learns encoded vector together with a posterior distribution, when pure PSA doesn't.

By comparing with a simple way of transfer learning called ``fine-tuning with special tokens'' (FIST) \cite{fang2021outline}, we investigate the effect of incorporating a latent code into the decoding. FIST does not learn a latent code, but only fine tunes a pre-trained GPT2 model with augmented language texts, {\it i.e.}, directly connecting prompt with story and put a special token `$<\!\!|\text{endoftext}|\!\!>$'' in between.

By comparing different ways of latent code injection, we evaluate their effectiveness accordingly. We label our CVAE model with the latent code injections \textcircled{1}, \textcircled{2} and \textcircled{3} as CVAE-\textcircled{1}, CVAE-\textcircled{2} and CVAE-\textcircled{3}, respectively, as is reflected in Figure~\ref{fig:tcvae}.

\subsection{Implementation Details}
We implement our models using the ``Huggingface Transformers'' library in Pytorch \cite{Wolf2019HuggingFacesTS}. In evaluation, we generate stories using the top-k top-p random sampling scheme \cite{holtzman2019curious,keskar2019ctrl} with $k=100$ and $p=0.9$. Temperature smoothing technique is also applied with $T=0.9$. Considering the two relatively large test datasets, we randomly decode one story per test input, rather than sampling several stories per test prompt and selecting the best one.

\subsection{Evaluation Results}

\begin{table*}[t]
\begin{centering}
\begin{adjustbox}{scale={1.1}{1.1},center}
\begin{tabular}{c|c|c|c|c|c|c|c|c|c|c|c}
\toprule
\multirow{2}{*}{Methods} & \multicolumn{2}{c|}{Perplexity\! $\downarrow$} & \multicolumn{3}{c|}{ROUGE-1\! $\uparrow$} & \multicolumn{3}{c|}{ROUGE-2\! $\uparrow$} & \multicolumn{3}{c}{ROUGE-L\! $\uparrow$} \tabularnewline
\cline{2-3}\cline{4-12}
 & Word & BPE & F1 & P & R & F1 & P & R & F1 & P & R \tabularnewline
\toprule
\multicolumn{12}{c}{Dataset: $\mathtt{WritingPrompts}$}\tabularnewline
\toprule
Fusion & 36.0 & - & 0.223 & \textbf{0.386} & 0.157 & 0.038 & 0.074 & 0.026 & 0.206 & \textbf{0.358} & 0.145 \tabularnewline
PSA & 31.6 & \textbf{21.3} & 0.265 & 0.316 & \textbf{0.228} & 0.047 & 0.054 & \textbf{0.041} & 0.248 & 0.296 & \textbf{0.213} \tabularnewline
FIST & 30.2 & 25.9 & 0.181 & 0.339 & 0.123 & 0.023 & 0.046 & 0.015 & 0.17 & 0.321 & 0.116 \tabularnewline
\midrule
CVAE \textcircled{1} & \textbf{26.4} & 23.2 & \textbf{0.265} & 0.332 & 0.221 & 0.046 & 0.067 & 0.035 & 0.244 & 0.353 & 0.187 \tabularnewline
CVAE \textcircled{2} & \textbf{26.8} & 23.2 & \textbf{0.266} & 0.325 & 0.225 & \textbf{0.049} & \textbf{0.074} & 0.037 & \textbf{0.253} & 0.348 & 0.199 \tabularnewline
\bottomrule
\multicolumn{12}{c}{Dataset: $\mathtt{WikiPlots}$}\tabularnewline
\toprule
Fusion & 108.2 & - & 0.185 & 0.185 & 0.185 & 0.026 & 0.027 & 0.025 & 0.15 & 0.149 & 0.151 \tabularnewline
PSA & 79.5 & 47.8 & 0.188 & 0.188 & \textbf{0.189} & 0.026 & 0.025 & \textbf{0.027} & 0.172 & 0.171 & 0.173 \tabularnewline
FIST & 38.9 & 26.5 & 0.166 & \textbf{0.253} & 0.124 & 0.018 & 0.032 & 0.013 & 0.15 & \textbf{0.231} & 0.111 \tabularnewline
\midrule
CVAE \textcircled{1} & \textbf{37.6} & \textbf{25.4} & \textbf{0.196} & 0.211 & 0.183 & \textbf{0.026} & 0.027 & 0.025 & \textbf{0.187} & 0.196 & \textbf{0.178} \tabularnewline
CVAE \textcircled{2} & \textbf{37.6} & \textbf{26.4} & \textbf{0.190} & 0.221 & 0.167 & \textbf{0.030} & \textbf{0.035} & 0.026 & 0.169 & 0.197 & 0.148 \tabularnewline
\bottomrule
\end{tabular}
\end{adjustbox}
\end{centering}
\caption{Automatic metrics for conditional story generation evaluated on two datasets. }
\label{table:automatic_metrics_tvae}
\end{table*}

\paragraph{Automatic Metrics:}
We evaluate the following automatic metrics towards target stories:
\begin{itemize}
\item Perplexity (PPL) is used to evaluate language models and often regarded as a proxy for generation quality. All models based on GPT-2 use the BPE tokenization scheme, where PPL values are not directly comparable with some previous models such as \cite{fan2018hierarchical} with PPLs computed at the natural word level. Similar to \cite{see2019massively}, we additionally compute the word-level perplexity of GPT-2 models to enable the comparison with previous models. That is, we normalize the total negative log probability of the target text by the number of word level tokens.
\item ROUGE scores are computed as n-gram overlap of test generated stories versus given target stories. For completeness, we report ROUGE scores (ROUGE-1, ROUGE-2, and ROUGE-L) \cite{lin2002manual} of n-gram overlap with both precision (P), recall (R), and F1.
\end{itemize}

\paragraph{Automatic Evaluation Results:}
The results are presented in Table~\ref{table:automatic_metrics_tvae}. 

Overall, our CVAE model achieves generally better, at least comparable, metrics in terms of lower PPL and higher ROUGE scores, demonstrating state-of-the-art conditional story generation performance. 

Methods based on pre-trained models (PSA / FIST / CVAE) show better overall performance with relatively less task specific efforts than the Fusion models, demonstrating the power of large-scale pre-training with Transformer-based architecture. Fusion models still show relatively high precision scores, especially in $\mathtt{WritingPrompts}$, due to its dedicated design for story generation.

When comparing CVAE \textcircled{2} with PSA, we observe performance improvement due to the flexible learned representation space. Note that CVAE merges a sequence of encoding representation vectors into a single latent vector, which is the key difference with original PSA.

When comparing CVAE variants with FIST, we observe the benefit of latent representation modeling as a powerful addition to pure occurance modeling. 

When comparing different ways of latent code injection in CVAE, we observe that it is hard to made option \textcircled{3} work empirically; options \textcircled{1} and \textcircled{2} perform comparably well.\footnote{We also observe that using both \textcircled{1} and \textcircled{2} does not consistently improve performance. } Our observation is different from \cite{li2020optimus}, which claims \textcircled{2} works significantly better than \textcircled{1}. We suspect this is due to an inherently different experiment setting, where we work with significantly longer text.

\begin{figure}[t!]
\centering
\includegraphics[scale=0.6]{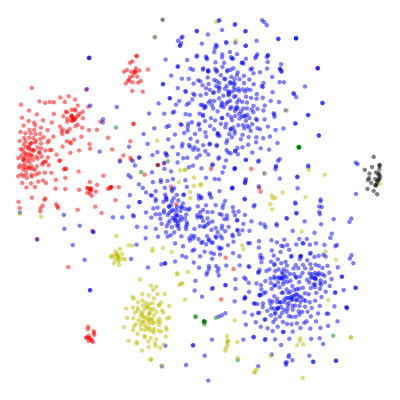}
\vspace{-0mm}
\caption{Visualization of prompts in $\mathtt{WikiPlots}$ test dataset through t-SNE. Item types are draw in different colors: red for TV series; blue for films; yellow for manga and comic; black for novel.}
\label{fig:repre-wi}
\end{figure}

\paragraph{Qualitative Evaluation:}
When conducting training on the $\mathtt{WikiPlots}$ dataset, we observe similar representation learning results as shown in Figure~\ref{fig:repre-wi}. A story prompt in $\mathtt{WikiPlots}$ may have extractable key words to reveal the item types, such as TV series, film, music, manga, novel, and game, {\it etc}. We observe that item types from test story prompts are clearly clustered in the latent code space, which implies effective representation learning to capture inherent characteristic of the prompts. 

We further present qualitative generation examples on the test datasets in Tables~\ref{table:example_wp}--\ref{table:example_wi_control}. We observe that stories are semantically and grammatically sound, and more importantly, highly conditioned on and consistent with given prompts. A large scale human evaluation is underway, which is quite expensive due to text length and evaluation scale.

\paragraph{Controllability Experiments:} 
To verify the effect of the learned latent representation vectors in generation, we conduct an interesting ``control'' experiment: given two prompt $\xv_1$ and $\xv_2$, we generate story from $p_{\thetav}(\yv|\xv_1, \zv_2)$, {\it i.e.}, conditioning on $\xv_1$ prefix and feeding in latent code $\zv_2$ along the decoding. In this way, the generated story will lie in the combination semantic space of the two prompt $\xv_1$ and $\xv_2$, especially after the latent code $\zv_2$ takes effect and dominates. 

Generation examples from the two test datasets are presented in Tables~\ref{table:example_wp_control} and \ref{table:example_wi_control}. We colorize key words in generated stories that coincides with given prompts $\xv_1$ and $\xv_2$ accordingly. Such examples confirm the effect of latent codes in generation, indicating our model as a principal way to enhance controllability.

\section{Conclusion and Future Research}
In this paper, we propose Transformer-based latent variable models to enhance story controllability while maintaining state-of-the-art generation effectiveness. Our test bed is a much more challenging and under-explored long-text application than the traditional short-text generation. Our results indicate the superiority of Transformer-based latent variable models, and appeal more efforts to be invested in the domain. 

\clearpage{} 


\bibliography{refs}

\begin{thebibliography}{48}
\providecommand{\natexlab}[1]{#1}
\providecommand{\url}[1]{\texttt{#1}}
\providecommand{\urlprefix}{URL }
\expandafter\ifx\csname urlstyle\endcsname\relax
  \providecommand{\doi}[1]{doi:\discretionary{}{}{}#1}\else
  \providecommand{\doi}{doi:\discretionary{}{}{}\begingroup
  \urlstyle{rm}\Url}\fi

\bibitem[{Bao et~al.(2019{\natexlab{a}})Bao, He, Wang, and Wu}]{bao2019plato}
Bao, S.; He, H.; Wang, F.; and Wu, H. 2019{\natexlab{a}}.
\newblock PLATO: Pre-trained Dialogue Generation Model with Discrete Latent
  Variable.
\newblock \emph{arXiv preprint arXiv:1910.07931} .

\bibitem[{Bao et~al.(2019{\natexlab{b}})Bao, Zhou, Huang, Li, Mou, Vechtomova,
  Dai, and Chen}]{bao2019generating}
Bao, Y.; Zhou, H.; Huang, S.; Li, L.; Mou, L.; Vechtomova, O.; Dai, X.; and
  Chen, J. 2019{\natexlab{b}}.
\newblock Generating Sentences from Disentangled Syntactic and Semantic Spaces.
\newblock In \emph{Proceedings of the 57th Annual Meeting of the Association
  for Computational Linguistics}, 6008--6019.

\bibitem[{Bourlard and Kamp(1988)}]{bourlard1988auto}
Bourlard, H.; and Kamp, Y. 1988.
\newblock Auto-association by multilayer perceptrons and singular value
  decomposition.
\newblock \emph{Biological cybernetics} 59(4-5): 291--294.

\bibitem[{Bowman et~al.(2015)Bowman, Vilnis, Vinyals, Dai, Jozefowicz, and
  Bengio}]{bowman2015generating}
Bowman, S.~R.; Vilnis, L.; Vinyals, O.; Dai, A.~M.; Jozefowicz, R.; and Bengio,
  S. 2015.
\newblock Generating sentences from a continuous space.
\newblock \emph{arXiv preprint arXiv:1511.06349} .

\bibitem[{Cheng et~al.(2019)Cheng, Xu, Wang, Chu, Huang, Chen, and
  Hu}]{cheng2019variational}
Cheng, X.; Xu, W.; Wang, T.; Chu, W.; Huang, W.; Chen, K.; and Hu, J. 2019.
\newblock Variational Semi-Supervised Aspect-Term Sentiment Analysis via
  Transformer.
\newblock In \emph{Proceedings of the 23rd Conference on Computational Natural
  Language Learning (CoNLL)}, 961--969.

\bibitem[{Cho et~al.(2014)Cho, van Merri{\"e}nboer, Gulcehre, Bahdanau,
  Bougares, Schwenk, and Bengio}]{cho2014learning}
Cho, K.; van Merri{\"e}nboer, B.; Gulcehre, C.; Bahdanau, D.; Bougares, F.;
  Schwenk, H.; and Bengio, Y. 2014.
\newblock Learning Phrase Representations using RNN Encoder--Decoder for
  Statistical Machine Translation.
\newblock In \emph{Proceedings of the 2014 Conference on Empirical Methods in
  Natural Language Processing (EMNLP)}, 1724--1734.

\bibitem[{Dathathri et~al.(2019)Dathathri, Madotto, Lan, Hung, Frank, Molino,
  Yosinski, and Liu}]{dathathri2019plug}
Dathathri, S.; Madotto, A.; Lan, J.; Hung, J.; Frank, E.; Molino, P.; Yosinski,
  J.; and Liu, R. 2019.
\newblock Plug and Play Language Models: A Simple Approach to Controlled Text
  Generation.
\newblock In \emph{International Conference on Learning Representations}.

\bibitem[{Devlin et~al.(2018)Devlin, Chang, Lee, and
  Toutanova}]{devlin2018bert}
Devlin, J.; Chang, M.-W.; Lee, K.; and Toutanova, K. 2018.
\newblock Bert: Pre-training of deep bidirectional transformers for language
  understanding.
\newblock \emph{arXiv preprint arXiv:1810.04805} .

\bibitem[{Fan, Lewis, and Dauphin(2018)}]{fan2018hierarchical}
Fan, A.; Lewis, M.; and Dauphin, Y. 2018.
\newblock Hierarchical Neural Story Generation.
\newblock In \emph{Proceedings of the 56th Annual Meeting of the Association
  for Computational Linguistics (Volume 1: Long Papers)}, 889--898.

\bibitem[{Fang et~al.(2019)Fang, Li, Gao, Dong, and Chen}]{fang2019implicit}
Fang, L.; Li, C.; Gao, J.; Dong, W.; and Chen, C. 2019.
\newblock Implicit Deep Latent Variable Models for Text Generation.
\newblock In \emph{Proceedings of the 2019 Conference on Empirical Methods in
  Natural Language Processing and the 9th International Joint Conference on
  Natural Language Processing (EMNLP-IJCNLP)}, 3937--3947.

\bibitem[{Fang et~al.(2021)Fang, Zeng, Liu, Bo, Dong, and
  Chen}]{fang2021outline}
Fang, L.; Zeng, T.; Liu, C.; Bo, L.; Dong, W.; and Chen, C. 2021.
\newblock Outline to Story: Fine-grained Controllable Story Generation from
  Cascaded Events.
\newblock \emph{arXiv preprint arXiv:2101.00822} .

\bibitem[{Fu et~al.(2019)Fu, Li, Liu, Gao, Celikyilmaz, and
  Carin}]{fu2019cyclical}
Fu, H.; Li, C.; Liu, X.; Gao, J.; Celikyilmaz, A.; and Carin, L. 2019.
\newblock Cyclical Annealing Schedule: A Simple Approach to Mitigating KL
  Vanishing.
\newblock \emph{NAACL} .

\bibitem[{Han et~al.(2020)Han, Meng, Wu, and Li}]{han2020non}
Han, Q.; Meng, Y.; Wu, F.; and Li, J. 2020.
\newblock Non-Autoregressive Neural Dialogue Generation.
\newblock \emph{arXiv preprint arXiv:2002.04250} .

\bibitem[{He et~al.(2018)He, Spokoyny, Neubig, and
  Berg-Kirkpatrick}]{he2018lagging}
He, J.; Spokoyny, D.; Neubig, G.; and Berg-Kirkpatrick, T. 2018.
\newblock Lagging Inference Networks and Posterior Collapse in Variational
  Autoencoders.
\newblock In \emph{International Conference on Learning Representations}.

\bibitem[{Hochreiter and Schmidhuber(1997)}]{hochreiter1997long}
Hochreiter, S.; and Schmidhuber, J. 1997.
\newblock Long short-term memory.
\newblock \emph{Neural computation} 9(8): 1735--1780.

\bibitem[{Holtzman et~al.(2019)Holtzman, Buys, Du, Forbes, and
  Choi}]{holtzman2019curious}
Holtzman, A.; Buys, J.; Du, L.; Forbes, M.; and Choi, Y. 2019.
\newblock The Curious Case of Neural Text Degeneration.
\newblock In \emph{International Conference on Learning Representations}.

\bibitem[{Hu et~al.(2017)Hu, Yang, Liang, Salakhutdinov, and
  Xing}]{hu2017toward}
Hu, Z.; Yang, Z.; Liang, X.; Salakhutdinov, R.; and Xing, E.~P. 2017.
\newblock Toward controlled generation of text.
\newblock In \emph{Proceedings of the 34th International Conference on Machine
  Learning-Volume 70}, 1587--1596. JMLR. org.

\bibitem[{Kasai et~al.(2020)Kasai, Cross, Ghazvininejad, and
  Gu}]{kasai2020parallel}
Kasai, J.; Cross, J.; Ghazvininejad, M.; and Gu, J. 2020.
\newblock Parallel Machine Translation with Disentangled Context Transformer.
\newblock \emph{arXiv preprint arXiv:2001.05136} .

\bibitem[{Keskar et~al.(2019)Keskar, McCann, Varshney, Xiong, and
  Socher}]{keskar2019ctrl}
Keskar, N.~S.; McCann, B.; Varshney, L.~R.; Xiong, C.; and Socher, R. 2019.
\newblock Ctrl: A conditional transformer language model for controllable
  generation.
\newblock \emph{arXiv preprint arXiv:1909.05858} .

\bibitem[{Kingma and Welling(2013)}]{kingma2013auto}
Kingma, D.~P.; and Welling, M. 2013.
\newblock Auto-encoding variational bayes.
\newblock \emph{arXiv preprint arXiv:1312.6114} .

\bibitem[{Li et~al.(2020)Li, Gao, Li, Li, Peng, Zhang, and Gao}]{li2020optimus}
Li, C.; Gao, X.; Li, Y.; Li, X.; Peng, B.; Zhang, Y.; and Gao, J. 2020.
\newblock Optimus: Organizing Sentences via Pre-trained Modeling of a Latent
  Space.
\newblock \emph{arXiv preprint arXiv:2004.04092} .

\bibitem[{Lin and Hovy(2002)}]{lin2002manual}
Lin, C.-Y.; and Hovy, E. 2002.
\newblock Manual and automatic evaluation of summaries.
\newblock In \emph{Proceedings of the ACL-02 Workshop on Automatic
  Summarization-Volume 4}, 45--51. Association for Computational Linguistics.

\bibitem[{Lin et~al.(2020)Lin, Winata, Xu, Liu, and Fung}]{lin2020variational}
Lin, Z.; Winata, G.~I.; Xu, P.; Liu, Z.; and Fung, P. 2020.
\newblock Variational Transformers for Diverse Response Generation.
\newblock \emph{arXiv preprint arXiv:2003.12738} .

\bibitem[{Ma et~al.(2019)Ma, Zhou, Li, Neubig, and Hovy}]{ma2019flowseq}
Ma, X.; Zhou, C.; Li, X.; Neubig, G.; and Hovy, E. 2019.
\newblock FlowSeq: Non-Autoregressive Conditional Sequence Generation with
  Generative Flow.
\newblock In \emph{Proceedings of the 2019 Conference on Empirical Methods in
  Natural Language Processing and the 9th International Joint Conference on
  Natural Language Processing (EMNLP-IJCNLP)}, 4273--4283.

\bibitem[{Maaten and Hinton(2008)}]{maaten2008visualizing}
Maaten, L. v.~d.; and Hinton, G. 2008.
\newblock Visualizing data using t-SNE.
\newblock \emph{Journal of machine learning research} 9(Nov): 2579--2605.

\bibitem[{Mao et~al.(2019)Mao, Majumder, McAuley, and
  Cottrell}]{mao2019improving}
Mao, H.~H.; Majumder, B.~P.; McAuley, J.; and Cottrell, G. 2019.
\newblock Improving Neural Story Generation by Targeted Common Sense Grounding.
\newblock In \emph{Proceedings of the 2019 Conference on Empirical Methods in
  Natural Language Processing and the 9th International Joint Conference on
  Natural Language Processing (EMNLP-IJCNLP)}, 5990--5995.

\bibitem[{Miao, Yu, and Blunsom(2016)}]{miao2016neural}
Miao, Y.; Yu, L.; and Blunsom, P. 2016.
\newblock Neural variational inference for text processing.
\newblock In \emph{International conference on machine learning}, 1727--1736.

\bibitem[{Mostafazadeh et~al.(2016)Mostafazadeh, Chambers, He, Parikh, Batra,
  Vanderwende, Kohli, and Allen}]{mostafazadeh2016corpus}
Mostafazadeh, N.; Chambers, N.; He, X.; Parikh, D.; Batra, D.; Vanderwende, L.;
  Kohli, P.; and Allen, J. 2016.
\newblock A corpus and evaluation framework for deeper understanding of
  commonsense stories.
\newblock \emph{arXiv preprint arXiv:1604.01696} .

\bibitem[{Peng et~al.(2018)Peng, Ghazvininejad, May, and
  Knight}]{peng2018towards}
Peng, N.; Ghazvininejad, M.; May, J.; and Knight, K. 2018.
\newblock Towards controllable story generation.
\newblock In \emph{Proceedings of the First Workshop on Storytelling}, 43--49.

\bibitem[{Radford et~al.(2018)Radford, Narasimhan, Salimans, and
  Sutskever}]{radford2018improving}
Radford, A.; Narasimhan, K.; Salimans, T.; and Sutskever, I. 2018.
\newblock Improving language understanding by generative pre-training.
\newblock \emph{URL https://s3-us-west-2. amazonaws.
  com/openai-assets/researchcovers/languageunsupervised/language understanding
  paper. pdf} .

\bibitem[{Radford et~al.(2019)Radford, Wu, Child, Luan, Amodei, and
  Sutskever}]{radford2019language}
Radford, A.; Wu, J.; Child, R.; Luan, D.; Amodei, D.; and Sutskever, I. 2019.
\newblock Language models are unsupervised multitask learners.
\newblock \emph{OpenAI Blog} 1(8): 9.

\bibitem[{Rezende, Mohamed, and Wierstra(2014)}]{rezende2014stochastic}
Rezende, D.~J.; Mohamed, S.; and Wierstra, D. 2014.
\newblock Stochastic Backpropagation and Approximate Inference in Deep
  Generative Models.
\newblock In \emph{International Conference on Machine Learning}, 1278--1286.

\bibitem[{See et~al.(2019)See, Pappu, Saxena, Yerukola, and
  Manning}]{see2019massively}
See, A.; Pappu, A.; Saxena, R.; Yerukola, A.; and Manning, C.~D. 2019.
\newblock Do Massively Pretrained Language Models Make Better Storytellers?
\newblock In \emph{Proceedings of the 23rd Conference on Computational Natural
  Language Learning (CoNLL)}, 843--861.

\bibitem[{Sergio(2019)}]{cunha_arxiv}
Sergio, G.~C. 2019.
\newblock ArXivAbsTitleDataset: Extracting Abstract and Title Dataset from
  arXiv articles.
\newblock \emph{Github repository} .

\bibitem[{Shah and Barber(2018)}]{shah_generative}
Shah, H.; and Barber, D. 2018.
\newblock Generative Neural Machine Translation.
\newblock In Bengio, S.; Wallach, H.; Larochelle, H.; Grauman, K.;
  Cesa-Bianchi, N.; and Garnett, R., eds., \emph{Advances in Neural Information
  Processing Systems 31}, 1346--1355. Curran Associates, Inc.
\newblock
  \urlprefix\url{http://papers.nips.cc/paper/7409-generative-neural-machine-translation.pdf}.

\bibitem[{Shen et~al.(2017)Shen, Lei, Barzilay, and Jaakkola}]{shen2017style}
Shen, T.; Lei, T.; Barzilay, R.; and Jaakkola, T. 2017.
\newblock Style transfer from non-parallel text by cross-alignment.
\newblock In \emph{Advances in neural information processing systems},
  6830--6841.

\bibitem[{Shu et~al.(2020)Shu, Lee, Nakayama, and Cho}]{shu2020latent}
Shu, R.; Lee, J.; Nakayama, H.; and Cho, K. 2020.
\newblock Latent-Variable Non-Autoregressive Neural Machine Translation with
  Deterministic Inference Using a Delta Posterior.
\newblock In \emph{AAAI}.

\bibitem[{Vaswani et~al.(2017)Vaswani, Shazeer, Parmar, Uszkoreit, Jones,
  Gomez, Kaiser, and Polosukhin}]{vaswani2017attention}
Vaswani, A.; Shazeer, N.; Parmar, N.; Uszkoreit, J.; Jones, L.; Gomez, A.~N.;
  Kaiser, {\L}.; and Polosukhin, I. 2017.
\newblock Attention is all you need.
\newblock In \emph{Advances in neural information processing systems},
  5998--6008.

\bibitem[{Wang and Wan(2019)}]{wang2019t}
Wang, T.; and Wan, X. 2019.
\newblock T-CVAE: Transformer-based conditioned variational autoencoder for
  story completion.
\newblock In \emph{Proceedings of the 28th International Joint Conference on
  Artificial Intelligence}, 5233--5239. AAAI Press.

\bibitem[{Wang et~al.(2020)Wang, Wang, An, Yu, and Chen}]{wang2020faithful}
Wang, Z.; Wang, X.; An, B.; Yu, D.; and Chen, C. 2020.
\newblock Towards Faithful Neural Table-to-Text Generation with
  Content-Matching Constraints.

\bibitem[{Wolf et~al.(2019)Wolf, Debut, Sanh, Chaumond, Delangue, Moi, Cistac,
  Rault, Louf, Funtowicz, and Brew}]{Wolf2019HuggingFacesTS}
Wolf, T.; Debut, L.; Sanh, V.; Chaumond, J.; Delangue, C.; Moi, A.; Cistac, P.;
  Rault, T.; Louf, R.; Funtowicz, M.; and Brew, J. 2019.
\newblock HuggingFace's Transformers: State-of-the-art Natural Language
  Processing.
\newblock \emph{ArXiv} abs/1910.03771.

\bibitem[{Yang et~al.(2017)Yang, Hu, Salakhutdinov, and
  Berg-Kirkpatrick}]{yang2017improved}
Yang, Z.; Hu, Z.; Salakhutdinov, R.; and Berg-Kirkpatrick, T. 2017.
\newblock Improved variational autoencoders for text modeling using dilated
  convolutions.
\newblock In \emph{Proceedings of the 34th International Conference on Machine
  Learning-Volume 70}, 3881--3890. JMLR. org.

\bibitem[{Yao et~al.(2019)Yao, Peng, Weischedel, Knight, Zhao, and
  Yan}]{yao2019plan}
Yao, L.; Peng, N.; Weischedel, R.; Knight, K.; Zhao, D.; and Yan, R. 2019.
\newblock Plan-and-write: Towards better automatic storytelling.
\newblock In \emph{Proceedings of the AAAI Conference on Artificial
  Intelligence}, volume~33, 7378--7385.

\bibitem[{Zhao et~al.(2018)Zhao, Kim, Zhang, Rush, and
  LeCun}]{zhao2018adversarially}
Zhao, J.; Kim, Y.; Zhang, K.; Rush, A.; and LeCun, Y. 2018.
\newblock Adversarially regularized autoencoders.
\newblock In \emph{International Conference on Machine Learning}, 5902--5911.
  PMLR.

\bibitem[{Zhao, Lee, and Eskenazi(2018)}]{zhao2018unsupervised}
Zhao, T.; Lee, K.; and Eskenazi, M. 2018.
\newblock Unsupervised Discrete Sentence Representation Learning for
  Interpretable Neural Dialog Generation.
\newblock In \emph{Proceedings of the 56th Annual Meeting of the Association
  for Computational Linguistics (Volume 1: Long Papers)}, 1098--1107.

\bibitem[{Zhao, Zhao, and Eskenazi(2017)}]{zhao2017learning}
Zhao, T.; Zhao, R.; and Eskenazi, M. 2017.
\newblock Learning Discourse-level Diversity for Neural Dialog Models using
  Conditional Variational Autoencoders.
\newblock In \emph{Proceedings of the 55th Annual Meeting of the Association
  for Computational Linguistics (Volume 1: Long Papers)}, 654--664.

\bibitem[{Zhou and Neubig(2017)}]{zhou2017multi}
Zhou, C.; and Neubig, G. 2017.
\newblock Multi-space variational encoder-decoders for semi-supervised labeled
  sequence transduction.
\newblock \emph{arXiv preprint arXiv:1704.01691} .

\bibitem[{Ziegler et~al.(2019)Ziegler, Melas-Kyriazi, Gehrmann, and
  Rush}]{ziegler2019encoder}
Ziegler, Z.~M.; Melas-Kyriazi, L.; Gehrmann, S.; and Rush, A.~M. 2019.
\newblock Encoder-Agnostic Adaptation for Conditional Language Generation.
\newblock \emph{arXiv preprint arXiv:1908.06938} .

\end{thebibliography}

\clearpage{} 

\onecolumn
\appendix
\begin{center}\LARGE{Appendix}\end{center}


\begin{table*}[htbp] \normalsize{}
\begin{tcolorbox}
\hspace{-7mm}
\begin{tabular}{l p{0.82\linewidth}}

{\bf Prompt:} & [ WP ] \textcolor{red}{The root password for the universe is discovered.} \\

{\bf \textit{Target Story}} & ------------------------------------------------------------------------------- \\

 & The \textcolor{red}{password} wasn't so simple. It required a sequence of well-timed thoughts, followed a series of moods, and then a frequency inaudible to the human ear. \\ \\

 & And yet, I cracked it. The possibilities opened up before me, vast and terrifying in all their permutations. The longer I stared into the truth, the reasons, the very secrets behind everything, the more I realized that neither I nor anyone else could properly harness the immediacy of all knowledge. \\ \\

 & Ignorance wasn't necessarily bliss, but there was a joy in gradual discovery. Humanity was meant to play this game without cheat codes. 
 I changed the \textcolor{red}{password} before wiping my memory clean. \\
\\
{\bf \textit{Generated Story}} & ------------------------------------------------------------------------------- \\
 & \textcolor{red}{The root password for the universe is discovered} \\ \\
 & For over two thousand years we have searched the \textcolor{red}{universe for the root password}. From the \textcolor{red}{universe}'s earliest stages we have always received the same answer. The \textcolor{red}{password} has been our forever since it was \textcolor{red}{discovered}. The \textcolor{red}{universe} has no known method of creation other than itself to try and destroy it. The \textcolor{red}{password} is based off of the last known version of it. The \textcolor{red}{universe} is expanding and expanding and expanding. The \textcolor{red}{universe} seems to be expanding towards a greater dimension. What does this mean to us? Why? The answer is simple to discover. I have learned from thousands of sources and I have learned to use the truth for what I wish to be the true \textcolor{red}{universe}. \\ \\
 & I have no idea what is happening, or where it is going. The \textcolor{red}{universe} seems to be expanding, changing, and expanding away from where it started. Now, I know it is a mistake, but the truth must be known. I would like to contact the \textcolor{red}{universe}. The answer must be obvious to everyone. The answer must \textcolor{red}{be discovered}. All of my attempts to stop this thing have failed, and now I know. My purpose is to destroy the \textcolor{red}{universe} and rid it of the \textcolor{red}{root password}.
\end{tabular}
\end{tcolorbox}
\caption{Generation examples on test dataset of $\mathtt{WritingPrompts}$.}
\label{table:example_wp}
\end{table*}

\begin{table*}[htbp] \small{}
\begin{tcolorbox}
\hspace{-7mm}
\begin{tabular}{l p{0.88\linewidth}}

{\bf Prompt:} & You're elected \textcolor{red}{president}, but you don't want to be president anymore, so you do ridiculous things to get \textcolor{red}{impeached}, but they all turn out to be the best decision for the country, making you more popular. \\

{\bf \textit{Target Story}} & ------------------------------------------------------------------------------- \\

 & I've tried everything. And they still seem to love me. I remember a time when the people hated me. That was a good time. When I first became \textcolor{red}{president}, I eliminated income tax to improve the economy and the people started a riot. So to spite them, and to get myself \textcolor{red}{impeached}, I initiated the Taco Tuesday initiative (free tacos every tuesday) and suddenly the people loved me. That boosted taco sales 180 \% and caused a boom in the economy. I tried everything to get back at them (requiring my cabinet members to wear costumes to work every day, insulting other countries leaders, holding witch hunts) but nothing has worked. Each time they come back praising me for my stupidity. During the last presidential address I blathered like an idiot, and they praised me for inventing a new language, on live television. It is now called American, and is being taught in schools. The time before I recited Adolf Hitler's Beer Hall Putsch speech, and they praised me for my original thinking. I don't understand these people. One week I say we shut down the government, they praise it. The next, I suggest we pay more to the government, and they praise it. A sane man would think that what I did was completely insane and would tank our nation, but everything I did made it better. Last night, I went completely naked on live television during my address and instead of getting condemned and arrested for indecent exposure, I was heralded as a hero for every group known to man. They called me a visionary for removing illusions of needs for clothing and material. We now have no industry for clothing, because everyone is nude, which in turn saved money on income for all of our citizens. I would normally think that the people are only following me to see how much I can humiliate myself, but everything I seem to do is actually improving the nation as a whole. I said we should power our cars with farts and scientists actually developed a fart powered car. Our dependency on foreign oil is dwindling. I've won more awards for the most stupid ideas. I have patents on Galchoopas, Kerflickdas, and Ladugas, and I don't even know what they do (all of them from my presidential address). I apparently described them in so much detail during my babbling that scientists had the devices ready by the next day. \\ \\

 & I only want to quit. The people are begging me to stay for another term. I'm going to begin shooting people at the address to get them to \textcolor{red}{impeach} me. I pray it happens. \\ \\

 & It turns out everyone I shot (all 30 of them) were all spies from other countries planning my assassination. Apparently the other countries were afraid of America becoming too strong so they all sent Assassins to take me out at the address. Now all of the countries leaders have to answer to us for trying to take me out. God help me I can't escape this. \\
\\
{\bf \textit{Generated Story}} & ------------------------------------------------------------------------------- \\
 & "Mama!" I cry. My mom is always so concerned with me. I have to take care of my mother. You know I hate her for making me make this decision, but the government needs me, its an extension of itself. The whole'free tacos every Tuesday' crap was a joke, I've been told to just use the taco Tuesday initiative as much as possible and save \$10 on food stamps. The fact that this stuff gets us is just plain laughable. The last time this happened, the \textcolor{red}{President} of the United States got sick from being in an abusive relationship with her secretary. The president asked for a taco Tuesday initiative, she needed to go on vacation to San Francisco to "help me out". This should have been the most fun part, but apparently, she wasn't listening to the President. Apparently she'd been drinking too much and got sick of being drunk while she was vacationing. So I put my beer in the fridge and filled the bowl. The whole thing was really just about the end of time for me. The President went home and I was woken up by the sound of footsteps coming up the stairs. The door to the beer hall putsch speech was open. I started laughing so hard. Before the President could finish saying, I got up and yelled "Hands up!" In that moment, I realized, my mom was probably one of those crazy aliens who just figured out a way to be president and became president. I was the most popular man in the country, but the President wasn't going to make me do this for her. "Daddy, are you okay?" she asked as soon as I stepped out of the bathroom. My mother hugged me and told me to get a better look for the President. Before I could, I reached out and started picking up the President's hair. There were two little holes all over it. I walked over and saw the tiny hole at the top of the head. I lifted the President's head over the hole and placed my hands on her head. She looked at me like I was insane, and I told her to go fuck herself. As I was thinking this, I looked at my hands and noticed that they were holding the President's shirt. I took out my phone and called 911. The man who killed the President stood there crying. He turned to me and said "Don't you ever think about putting this on."
\end{tabular}
\end{tcolorbox}
\caption{More generation examples on test dataset of $\mathtt{WritingPrompts}$.}
\label{table:example_wp_1}
\end{table*}

\begin{table*}[htbp] \normalsize{}
\begin{tcolorbox}
\hspace{-7mm}
\begin{tabular}{l p{0.82\linewidth}}

{\bf Prompt:} & Humanity is the most powerful race in the \textcolor{red}{galaxy}. Every other race wants to win our respect. Even for ordinary humans, it's become impossible to go off-world because our presence demands so much respect, awe, and asskissing. \\

{\bf \textit{Target Story}} & ------------------------------------------------------------------------------- \\

 & At first nobody knew what to do. When we found the other civilizations, I remember I was in the pentagon with the country's best scientists trying to unfold the events that had happened hours before when Dr.Greg Darkasian found the first signs of alien civilizations. That day no body had any answers. That was ten years ago, now we know where we fall on the galactic hierarchy, and much to the pleasure of Earths inhabitants, it's sitting right at the top. Today I work as an alien Slave trader, you see once we found the dozens of other societies we focused on controlling them and the business that was booming was alien slave trade. So as a struggling scientist I turned to what humans do best. Enslaving our inferiors, much like societies have done for ages. Not only did we dominate these societies but we ravaged all of their home planets of any resources and now all that's useful from them is their physical labor. Of the few alien planets left untouched, they are weak and we could probably take them out with a single battalion. The only thing we didn't expect is that when these resources ran dry and the slaves mainly dead, our own planet was In turmoil. You see while all the countries of earth we're focusing on interplanetary domination. We failed to notice the growth of Japan, the one country that was not sucked Into the Galactic power race. Now japan threatens to control our home country and there's nothing we can do. Most of the worlds military services are deployed eons away from earth. \\
\\
{\bf \textit{Generated Story}} & ------------------------------------------------------------------------------- \\
 & We’ re on the brink of war, hundreds of worlds military services are gone, our allies are gone, some are captured, most are executed, we’ ve had ten years ago our only chance to stop it is to pull the trigger. It has been a month, we’ re being attacked on all fronts and our numbers are growing exponentially. We’ re sending troops to every sector of our \textcolor{red}{galaxy}, we’ re training for battle, we’ ve got weapons we need to fend off an alien invasion. We’ re sending reinforcements, the first battle with it was ten years ago. They came out of nowhere, they’ re huge, they’ re faster, they can fly, and they don’ t fight like us, but they fight on, because they don’ t have any of the same skills as us. All because of their stupid species. We’ re the biggest threat to them, we’ re the ones on the front lines. And the only way to defeat them is to kill them and replace them with the ones we have, but not because our stupid species wouldn’ t stand a chance, not because it would be much easier to defeat them, but because they’ re such an ignorant race. We’ re the only ones on the front lines, we’ re the only ones who don’ t believe that we need to kill all the other races. I don’ t know if there is any other species out there, there are only us. If there was any other race, we would be the ones on the front lines. The only ones that would take it, the only ones who would stand against us. Not even them, but them, because of our stupidity, because they don’ t have the brains to fight us on this side. I feel that if we all die here we’ ll be nothing more than shells and the planet is doomed to implode. Just like any other race, the only thing to do is to go with them, destroy the entire planet, and we won’ t be there, we’ ll be gone, the only other species on this planet, we’ ll be the only race alive, except for us. I don’ t know if it is better for them to simply leave us, we’ ll be dead, they’ re just the ones that are killing themselves. I wish they would stop us, they’ d kill us, they’ re not even smart enough to put up a fight, we’ ll be dead, the only species in this universe that has left, is us.
\end{tabular}
\end{tcolorbox}
\caption{More generation examples on test dataset of $\mathtt{WritingPrompts}$.}
\label{table:example_wp_2}
\end{table*}


\begin{table*}[htbp] \normalsize{}
\begin{tcolorbox}
\hspace{-7mm}
\begin{tabular}{l p{0.82\linewidth}}

{\bf Prompt 1:} & [ WP ] \textcolor{red}{The root password for the universe is discovered.} \\

{\bf \textit{Target Story}} & ------------------------------------------------------------------------------- \\

 & The \textcolor{red}{password} wasn't so simple. It required a sequence of well-timed thoughts, followed a series of moods, and then a frequency inaudible to the human ear. \\ \\

 & And yet, I cracked it. The possibilities opened up before me, vast and terrifying in all their permutations. The longer I stared into the truth, the reasons, the very secrets behind everything, the more I realized that neither I nor anyone else could properly harness the immediacy of all knowledge. \\ \\

 & Ignorance wasn't necessarily bliss, but there was a joy in gradual discovery. Humanity was meant to play this game without cheat codes. 
 I changed the \textcolor{red}{password} before wiping my memory clean. \\
\\
{\bf Prompt 2:} & [ WP ] Long Ago, \textcolor{blue}{God} sent his son Jesus to Earth. Now, Satan sends his daughter, Jessica... \\
{\bf \textit{Generated Story}} & ------------------------------------------------------------------------------- \\
 & \textcolor{red}{**The Root password for the universe is discovered. **} \\ \\
 & So I think my story really is unique because when I first started writing, it was a little hard to work up the courage to ask what it was that made the \textcolor{red}{universe} happen. I don ’ t know why, but I was trying to figure out what it really is. \\ \\
 & I guess it was something I heard in school. I ’ ve heard it all my life: how it is the real reason we are being built into the world. \\ \\
 & “ I’ m \textcolor{blue}{God}. ” I told myself as I had a hard time believing it. \\ \\
 & Anyway, I ’ m a \textcolor{blue}{god} now. I am alone and I believe in myself because I know the \textcolor{red}{universe} is about to change.
\end{tabular}
\end{tcolorbox}
\caption{Generation examples on test dataset of $\mathtt{WritingPrompts}$ with two prompts in control experiment.}
\label{table:example_wp_control}
\end{table*}


\begin{table*}[htbp] \normalsize{}
\begin{tcolorbox}
\hspace{-7mm}
\begin{tabular}{l p{0.9\linewidth}}

{\bf Prompt:} & \textcolor{red}{Bloodthirsty}: One Nation \textcolor{red}{Under Water} (novel) \\

{\bf \textit{Target Story}} & ------------------------------------------------------------------------------- \\

 & The story begins with Coast Guard rescue swimmer Virgil LaFleur as he rescues victims of Hurricane Katrina from their rooftops in New Orleans in the aftermath of the infamous flood of 2005. When Virgil's younger brother, Trey, informs him that their parents have not checked in at the evacuee station at Louis Armstrong airport, Virgil gets his Coast Guard pilot to take him to his parents' flooded neighborhood. There, Virgil swims through his parents' submerged home to find his parents stuck in their attic - his father already deceased and his mother near death. Virgil attempts to swim his mother out through the flooded house, but nearly drowns and is pulled from the \textcolor{red}{water} by Trey and another Coast Guard crew member. \\ \\

 & Ten years later, the comic recaps that Virgil couldn't go near the \textcolor{red}{water} after the incident, and is now working as a ditch digger after having put his younger brother Trey through college (fulfilling a promise he made to his dying mother). A new storm - Rose - is on the horizon, and Virgil is planning to evacuate New Orleans forever. He begs Trey to go with him, but Trey - now a scientist - is in the middle of an exciting - and secretive - longevity project at Wolfinger Biomed, one of many companies owned by local carpetbagger Simon Wolfinger. Wolfinger is portrayed as a disaster capitalist who - along with partner-in-crime Howard Lawrence - preyed upon the city's weakened post-disaster status to enrich themselves through greed and corruption. When Trey dies in a fire at Wolfinger Biomed, Virgil has reason to suspect that his brother was murdered. \\ \\

 & Having nothing left to live for, Virgil decides to investigate his brother's death, even if that means staying in New Orleans long enough to die in the coming hurricane. What he uncovers is an \textcolor{red}{underground} cabal of hemovores - "an organism that ingests \textcolor{red}{blood} as the main part of its diet". Unlike vampires, the hemovores in "\textcolor{red}{Bloodthirsty}" are not supernatural beings; they are humans with a mutated gene that controls longevity. Their bodies are in a constant state of repair, which deprives their cells of oxygen. In order to compensate for this loss, they must consume fresh (oxygenated) human \textcolor{red}{blood} daily or they will die. Virgil learns that Trey had unwittingly taken part in a plan to synthesize the hemovore mutation, creating a serum that Wolfinger intends to sell on the black market to anyone wealthy enough to procure its fountain-of-youth-like properties. With only hours left before Hurricane Rose makes landfall in New Orleans, Virgil faces an uphill battle to stop Wolfinger from spreading the serum and killing thousands of innocent people in the process. \\
{\bf \textit{Generation}} & ------------------------------------------------------------------------------- \\
 & While trying to avoid shooting down a submarine that sank \textcolor{red}{under water}, a boy from that small school appears and explains the situation to the boy. A brief fight ensues, which is eventually led to the boy's being eaten by an insect-like creature. The boy then kills it and takes refuge in the same submarine that sank, only to have it reanimated.\\ \\
 & While the boy is unconscious he is taken to a veterinary hospital, where he attempts to convince the doctor that he is a reptile, but he is unable to talk. A biologist comes to his rescue, but he is unable to see anything. He goes to the psychiatrist, who informs him that the specimen of some unknown species is in the aquarium, and that he is suffering from a mental health problem. \\ \\
 & After a short wait in the hospital, the doctor is able to draw a connection between the creature he saw earlier and the creature he saw earlier, and they begin to build their own \textcolor{red}{underwater} lab. The doctor is told that he must take the creature back to the submarine, and so he decides to use the creature's memories to locate the creature, but only to find that it is not actually named. After the creature rescues the boy, the biologist helps him to open the laboratory and create a chamber filled with spores and insects. \\ \\
 & As the spores grow, the biologist states that he is making a discovery that will change his body forever and this is a revelation to the boy. He goes to another lab, and the biologist attempts to convince him to kill the creature. When the biologist tells him that he cannot kill him because of his size, the scientist agrees to euthanize him. 
\end{tabular}
\end{tcolorbox}
\caption{Generation examples on test dataset of $\mathtt{WikiPlots}$.}
\label{table:example_wi}
\end{table*}

\begin{table*}[htbp] \normalsize{}
\begin{tcolorbox}
\hspace{-7mm}
\begin{tabular}{l p{0.9\linewidth}}

{\bf Prompt:} & \textcolor{red}{Writers} (TV series) \\

{\bf \textit{Target Story}} & ------------------------------------------------------------------------------- \\

 & Emmett Shelby is a struggling young \textcolor{red}{writer} who, during a bout of writer's block, encounters fellow writer Jess Spencer who is suffering from the same condition. The two spend a day together, subsequently inspiring Emmett to continue writing. A year later, Emmett's successful twin sister Daisy is promoting her latest children's book when she runs into Jess, who happens to be an old school friend of hers. Daisy invites Jess to move in with her, setting the events of the series in motion. Daisy shares her home with two others - her boyfriend Barney Fields, and flamboyant landlord Hugh Darvill. Barney is an illustrator, providing the artwork to Daisy's bestselling children's novels. Hugh, on the other hand, is a failed erotic novelist, specialising in gay erotica aimed at children. Emmett still lives with his wealthy, controlling parents - whose grasp Daisy was able to escape some years prior - but ends up moving in with the others when they throw him out in the second episode of season one. Outside of the central group is Jenny Sinclair, Daisy's eccentric agent who bares an intense unrequited attraction towards Emmett. \\ \\

 & In the second season, Parker James III is introduced as an old friend of Daisy and Emmett's, and a new romantic interest for Jenny. Although the group are different, they quickly become close friends, and something akin to a family. The central arc of the series revolves around the budding romantic relationship between Jess and Emmett, which is hindered both by the arrival of Emmett's irritating girlfriend Clarice and Emmett's own cluelessness to his feelings. After sleeping together in the season one finale, Jess and Emmett finally begin a relationship together. \\
{\bf \textit{Generation}} & ------------------------------------------------------------------------------- \\
 & Set in London in the 1930s, the story follows the adventures of the main character, Daniel, as he tries to make his way in England from London to the East End, learning to write, with his flamboyant landlord Hugh Darvill. Despite the stress that he will have to cope with alone, Daniel is very determined to change his life. His old school friend, gay Erotica aimed to write a history of the city in which gay characters lived their lives, but Daniel finds himself falling for Erotica. Initially, Daniel is determined to return to England to write an article on socialite Christina, but eventually, he chooses to go back to her for financial reasons and finds a job in the English dubbing service in England. When Daniel's brother Peter moves in next door to provide a flat for Daniel, Peter is intrigued by his idea. Together they are initially attracted to each other, but the love that develops between them is also very strong. As Daniel looks for a new flat, he gets a chance to get into an old school, but finds it to be a run-down, poor working-class establishment. There he meets and befriends two young women, lesbian Jenny, and lesbian Jenny's best friend, lesbian Rose, who become his new lovers, and their flamboyant boyfriend, gay Erotica. \\ \\
 & Jenny and Rose were sent to England as a group to \textcolor{red}{write} an article in the newspaper, but due to the anti-homosexual atmosphere that developed there, they were sent to live in a local brothel. The series revolves around their stories and Daniel's attempts to get involved in their lives. In one episode, Jenny has to take a class in the arts because of her dislike of acting in public and her dislike of acting in a private capacity, while Daniel does not seem to find her suitable. The love between Jenny and Jenny's boyfriend, Parker James III, gradually builds to a point where they both become close friends.
\end{tabular}
\end{tcolorbox}
\caption{More generation examples on test dataset of $\mathtt{WikiPlots}$.}
\label{table:example_wi_1}
\end{table*}

\begin{table*}[htbp] \normalsize{}
\begin{tcolorbox}
\hspace{-7mm}
\begin{tabular}{l p{0.9\linewidth}}

{\bf Prompt:} & Mrittupuri: \textcolor{red}{Kill} Zone \\

{\bf \textit{Target Story}} & ------------------------------------------------------------------------------- \\

 & Sydney is a dream city for many immigrants, but also it is an ideal place for many international crime lords. Gangsters from Bangladesh, China, Lebanon, Russia have all moved to Australia, to take a piece of the pie, and try to build empire. Among them is the rising Bangladeshi gang Double Rajan Rahman (Rahmatullah), the kingpin of Bangladeshi gang is not happy with the way his elder son, Rajib Rahman (Saifullah Sadi) and his crime associate Raja (Arifin Shuvoo), run the business. The situation comes to a head after the timely death of Irish Mob boss’ son Jimmy, which prompts Rajib to leave the country. Raja maintains low profile during this period, and in the process falls in love with bar dancer Layla (Prosun Azad). Upon Rajib’s return, he warns Raja not to get involved with Layla, as he perceives Layla to be his own property. \\ \\

 & Raja manages to divert Rajib’s mind away, with plans to do business with Lengra Kashem (Taskeen Rahman), a mercenary from Bangladesh. Rajib and Sabbir, his youngest brother, meet Lengra and promise him the East coast territory, leading to their business partnership to become a booming new empire. \\
 
  & Eventually, Rahman and his crew fall through on their deal, and remove Lengra from the partnership. Lengra decides to avenge this deception, waiting for the right time to strike. \\
{\bf \textit{Generation}} & ------------------------------------------------------------------------------- \\
 & Mridupuri is set at a very dark time. The plot revolves around Karan (Arifin Shuvoo), run by his boss, Ravi Chatterjee, an underworld don in Mumbai. His boss is a serial \textcolor{red}{killer} who abducts the daughter of his colleague, Sivarasu (Baba Menon) in order to acquire more money for his wife's property in a jewel-like compound in Madras. After killing a lot of people in Madras, the mother of the kidnapped daughter is killed and the daughter is dragged to the same prison. In the prison the killer enters and kills all the guards while Karan escapes. It is revealed that the killer has come to kill Karan because of a personal vendetta and Karan wants to kill Karan for his daughter's sake. Karan is now very wealthy and wants to make a big profit in the diamond business and will soon be able to marry Sivarasu, a daughter of the murdered boss. Ravi Chatterjee, the crime associate Raja (Ajay Devgan) and his partner Shekhar (Lalitha Rajendran) are working hard to find the identity of the killer who is taking the daughter of the missing girl and use it against him. It is revealed that Karan's father, Ashokan (Arifin Shakuntala), is an underworld don who is working for the police. Raja is a well-respected and respected don who would like to marry Sivarasu so he would not kill Karan. Sivarasu wants to marry Sivarasu who is now in love with Ashokan and his fiancee Radha (Meena Devi) has already agreed to that. Raja maintains low profile of the murders and Karan does not get involved in it. The film ends with Sivarasu and Ravi Chatterjee in jail together.
\end{tabular}
\end{tcolorbox}
\caption{More generation examples on test dataset of $\mathtt{WikiPlots}$.}
\label{table:example_wi_2}
\end{table*}


\begin{table*}[htbp] \normalsize{}
\begin{tcolorbox}
\hspace{-7mm}
\begin{tabular}{l p{0.86\linewidth}}
{\bf Prompt 1:} & \textcolor{red}{Bloodthirsty}: One Nation \textcolor{red}{Under Water} (novel) \\

{\bf \textit{Target Story}} & ------------------------------------------------------------------------------- \\

 & The story begins with Coast Guard rescue swimmer Virgil LaFleur as he rescues victims of Hurricane Katrina from their rooftops in New Orleans in the aftermath of the infamous flood of 2005. When Virgil's younger brother, Trey, informs him that their parents have not checked in at the evacuee station at Louis Armstrong airport, Virgil gets his Coast Guard pilot to take him to his parents' flooded neighborhood. There, Virgil swims through his parents' submerged home to find his parents stuck in their attic - his father already deceased and his mother near death. Virgil attempts to swim his mother out through the flooded house, but nearly drowns and is pulled from the \textcolor{red}{water} by Trey and another Coast Guard crew member. \\ \\

 & Ten years later, the comic recaps that Virgil couldn't go near the \textcolor{red}{water} after the incident, and is now working as a ditch digger after having put his younger brother Trey through college (fulfilling a promise he made to his dying mother). A new storm - Rose - is on the horizon, and Virgil is planning to evacuate New Orleans forever. He begs Trey to go with him, but Trey - now a scientist - is in the middle of an exciting - and secretive - longevity project at Wolfinger Biomed, one of many companies owned by local carpetbagger Simon Wolfinger. Wolfinger is portrayed as a disaster capitalist who - along with partner-in-crime Howard Lawrence - preyed upon the city's weakened post-disaster status to enrich themselves through greed and corruption. When Trey dies in a fire at Wolfinger Biomed, Virgil has reason to suspect that his brother was murdered. \\ \\

 & Having nothing left to live for, Virgil decides to investigate his brother's death, even if that means staying in New Orleans long enough to die in the coming hurricane. What he uncovers is an \textcolor{red}{underground} cabal of hemovores - "an organism that ingests \textcolor{red}{blood} as the main part of its diet". Unlike vampires, the hemovores in "\textcolor{red}{Bloodthirsty}" are not supernatural beings; they are humans with a mutated gene that controls longevity. Their bodies are in a constant state of repair, which deprives their cells of oxygen. In order to compensate for this loss, they must consume fresh (oxygenated) human \textcolor{red}{blood} daily or they will die. Virgil learns that Trey had unwittingly taken part in a plan to synthesize the hemovore mutation, creating a serum that Wolfinger intends to sell on the black market to anyone wealthy enough to procure its fountain-of-youth-like properties. With only hours left before Hurricane Rose makes landfall in New Orleans, Virgil faces an uphill battle to stop Wolfinger from spreading the serum and killing thousands of innocent people in the process. \\
\\
{\bf Prompt 2:} & Queen of the \textcolor{blue}{Sea} (\textcolor{blue}{film}) \\
{\bf \textit{Generated Story}} & ------------------------------------------------------------------------------- \\
 & The \textcolor{blue}{film} tells the story of three high school students, James, his wife, and a 15-year-old girl, Lisa, at the end of the \textcolor{blue}{film}, who run into the \textcolor{blue}{sea} in an old fishing boat. They are captured by a fishing boat, who sends them \textcolor{red}{underwater}, with the intention of causing the boat to fall. The parents soon come to understand that the school swimmer may be the true killer, and so decide to bring him to school. \\ \\
 & The teacher, a teacher who has worked on the case for fifteen years, recognizes the fish and, on arriving, tries to convince him to let him swim. However, when it reaches the beach it's discovered that it's not Lisa. At first it seems to just be a fish. But when Lisa eventually notices it, she realizes that it is actually a human. Eventually, the fish becomes an outcast, and the girl, Lisa, and James start to try to convince them to let it swim. James gets drunk and runs out of the boat and finds himself covered in mud. He and Lisa are rescued by the teacher, who tells them that they need to swim for a chance at saving the girl's life.
\end{tabular}
\end{tcolorbox}
\caption{Generation examples on test dataset of $\mathtt{WikiPlots}$ with two prompts in control experiment.}
\label{table:example_wi_control}
\end{table*}


\end{document}